# Prior Guided Deep Difference Meta-Learner for Fast Adaptation to Stylized Segmentation

Anjali Balagopal, Dan Nguyen, Ti Bai, Michael Dohopolski, Mu-Han Lin, Steve Jiang

Medical Artificial Intelligence and Automation (MAIA) Laboratory and Department of Radiation Oncology, University of Texas Southwestern Medical Center, Dallas, Texas, USA

Email: Steve.Jiang@UTSouthwestern.edu

## ABSTRACT

Cancer radiotherapy treatment planning posts a unique image segmentation problem, where a particular anatomical structure may not be segmented simply based on its anatomical definition, rather, it may be segmented to facilitate the treatment plan optimization in very different ways by clinicians at different institutions. The different segmentation styles are typically based on institutional guidelines, clinical trial protocols, individual physicians' preference or specific dose planning requirements. Deep learning-based auto-segmentation models are often trained to segment organs according to their anatomical definitions. When such pre-trained general models are deployed at a new institution for radiotherapy treatment planning, the predicted organ contours may significantly differ from the segmentation styles of the local clinicians. Demanding a model adaptation may be challenging for some local clinical users without needed resources and deep learning expertise. To minimize the effort for clinical users to adapt a pre-trained auto-segmentation model to their own practice styles, we propose a Prior guided Deep Difference Meta-Learner (Prior-guided DDL).

When a pre-trained general auto-segmentation model is deployed at a new institution, a support framework in the proposed Prior-guided DDL network will learn the systematic difference between the model predictions and the final contours revised and approved by clinicians for an initial group of patients. The learned style feature differences are concatenated with the new patient's (query) features and then decoded to get the style-adapted segmentations. The model is independent of practice styles and anatomical structures. It meta-learns with simulated style differences and does not need to be exposed to any real clinical stylized structures during training. Once trained on the simulated data, it can be deployed for clinical use to adapt to new practice styles and new anatomical structures without further training.

To show the proof of concept, we tested the Prior-guided DDL network on six different practice style variations for three different anatomical structures. Pre-trained segmentation models were adapted from post-operative clinical target volume (CTV) segmentation to segment $CTV_{style1}$, $CTV_{style2}$, and $CTV_{style3}$, from parotid gland segmentation to segment $Parotid_{superficial}$, and from rectum segmentation to segment $Rectum_{superior}$ and $Rectum_{posterior}$. The mode performance was quantified with Dice Similarity Coefficient (DSC). With adaptation based on only first three patients, the average DSCs (%) were improved from 78.6, 71.9, 63.0, 52.2, 46.3 and 69.6 to 84.4, 77.8, 73.0, 77.8, 70.5, 68.1, for $CTV_{style1}$, $CTV_{style2}$, and $CTV_{style3}$, $Parotid_{superficial}$, $Rectum_{superior}$, and $Rectum_{posterior}$, respectively, showing the great potential of the Prior-guided DDL network for a fast and effortless adaptation to new practice styles.

## I. INTRODUCTION

The goal of radiotherapy is to achieve a full-dose coverage of the radiation target while sparing the dose to the organs-at-risk (OARs) nearby. The identification and the delineation of the target volumes and the OARs are critical to the quality of the radiation treatment planning and the safety of patient's treatment. As a standard of practice, physician will delineate and prescribe the intended doses to the target area as well as the tolerance dose to the OARs for the radiation treatment planners to create the optimal treatment plans. Hence, the contours reflect not only the notation of the anatomy but also the description of the desired dose objectives for the treatment.

Automatic segmentation for auto-contouring has drawn enormous attention in radiotherapy since it reduces the contouring time drastically and creates contours with less intra- and interobserver variability. There are numerous works of OAR segmentation proposed for radiotherapy treatment planning of different body sites. Atlas-based methods are among the most commonly used traditional approaches [1-8]. The optimal transformation between the atlas, which has the pre-segmented annotation map, and the image to be segmented is aligned by affine and deformable registration. Then the segmentation for the target image can be obtained by applying this transformation on the annotation map of the reference image. The reference images can be multiple ones with expert annotations or templates generated from the training set. Although atlas-based methods have the advantages of robustness and can perform segmentation without user interaction, they are based on image registration techniques and might generate incorrect organ maps if the organs are occupied by tumors. The time cost can be up to tens of minutes due to its extensive amount of computation.

Recently, convolutional neural networks (CNN) were adopted to advance structure delineation accuracy significantly. Ibragimov and Xing [10] proposed the first deep learning-based algorithm for radiotherapy OAR segmentation. They first detect OARs by automatically detecting the patient's head, which is used as a reference point for approximation of OARs positions and then train a patch-based CNN to classify voxels in the region of interest. Ren et al. [12] proposed an interleaved 3D-CNN for the joint segmentation of small organs in head and neck (H&N) region, where the region of interest is obtained via registration techniques. Following this, Zhu et al. [13], Tong et al. [14] , Tang et al. [15] also proposed advanced H&N segmentation models for achieving better segmentation performance. Many groups have designed novel deep CNN architectures for pelvic organ segmentation as well, [16,17,18]. Several of these papers have focused specifically on preoperative prostate gross tumor volume (GTV) and OARs [19,20,21,22,23] and a few on post-operative prostate clinical target volume (CTV) and OARS [24,25]. Even though these models are quite accurate, deploying these models into clinics comes with a lot of challenges.

In this work, we deal with one such challenge – labeling style variations, which is not a commonly researched topic albeit it being a very common barrier of deploying segmentation model in radiotherapy field. In radiotherapy, common structures segmented are GTV, CTV and OARs. GTV is the palpable disease that has to be treated with radiation. Using their knowledge of the disease, physicians then expand the GTV to create the CTV, which includes the microscopic extensions invisible in images. The CTV to be segmented is based on existing contouring guidelines and physicians' experience for the desired trade-off between OAR toxicity and tumor control, often resulting in labeling variations in CTV segmentation within an institution and across institutions [26,27]. Eliminating variations is neither possible nor desirable because CTV definition involves judgement around many different variables; there is not a unique correct CTV for a given tumor and there is no clinical trial proving that one definition is absolutely better than others. From many studies for many different cancer types, the magnitude of variations in contouring is known to be substantial [31].

When deploying a model outside the institution where it was developed, the segmentation style for CTV could be different depending on the specific institutional guidelines. In addition, OAR contours are often not segmented based on the anatomy definition as a full organ, but the functional definition that can better describe the desired spatial dose distribution for treatment planning purpose. A typical example is the contour of parotid gland for H&N cancer treatment planning. Instead of sparing the entire parotid gland, which could overlap with tumor and contradict with the goal of tumor coverage, sparing a single lobe of parotid that is further away from the tumor may achieve the same goal of parotid toxicity control without compromising the tumor coverage [28,29]. Another similar example is the posterior/superior aspect of rectum in the prostate cancer treatment planning [30].

To adapt a pre-trained general auto-segmentation model to a new institution with special segmentation styles, the local clinical users are required to have some knowledge on model adaptation or model training in addition to data acquisition and curation. Since difference between each special segmentation style and the segmentation of anatomical definition is quite consistent, we hypothesize that this difference can be learned from first several patients and then added to the pre-trained general model to produce a more precise "stylized' segmentation. By doing so, the local clinical users are only required to correct the results from the general model for the first several patients, without being required to collect and curate data and then explicitly adapt the model.

In this work, we propose a prior-guided Deep Difference Meta-Learner for Precise Stylized Segmentation (Prior-guided DDL) which is capable of quickly learning the difference between the results of a general segmentation model and a special segmentation style, and thus converting the auto-segmented labels to a desired style. We leverage systematic errors in the predictions from the pre-trained general segmentation models trained to generate new stylized segmentations. When distinguishable systematic segmentation style differences exist among different clinical users for an anatomical structure, typically for radiotherapy treatment planning purposes, it would be possible for our proposed Prior-guided DDL to learn these differences by simply using the pre-trained general model to several patients.

The clinical workflow for using the proposed Prior-guided DDL to a completely new segmentation style is illustrated in **Figure 1**. First, a pre-trained general auto-segmentation model generates the initial contour ($M_{1,new}$), and the local clinician reviews and manually corrects it to reach the new style ($P_{1,new}$). This corrected patient would now serve as the first prior patient. From the next patient onwards, the initial contour predicted by the pre-trained general auto-segmentation model is first corrected by the DDL network to generate $P_{2,new}$ with the help of $M_{1,new}$ and $P_{1,new}$, and then $P_{2,new}$ is reviewed and further corrected by the clinician if needed. The final contour $P_{2,new}$ along with $M_{2,new}$ serves as the second set of prior guidance. This process is repeated with more patients. At the stage when the model acceptably adapts to the new style, the correction required by the clinician would become minimal.

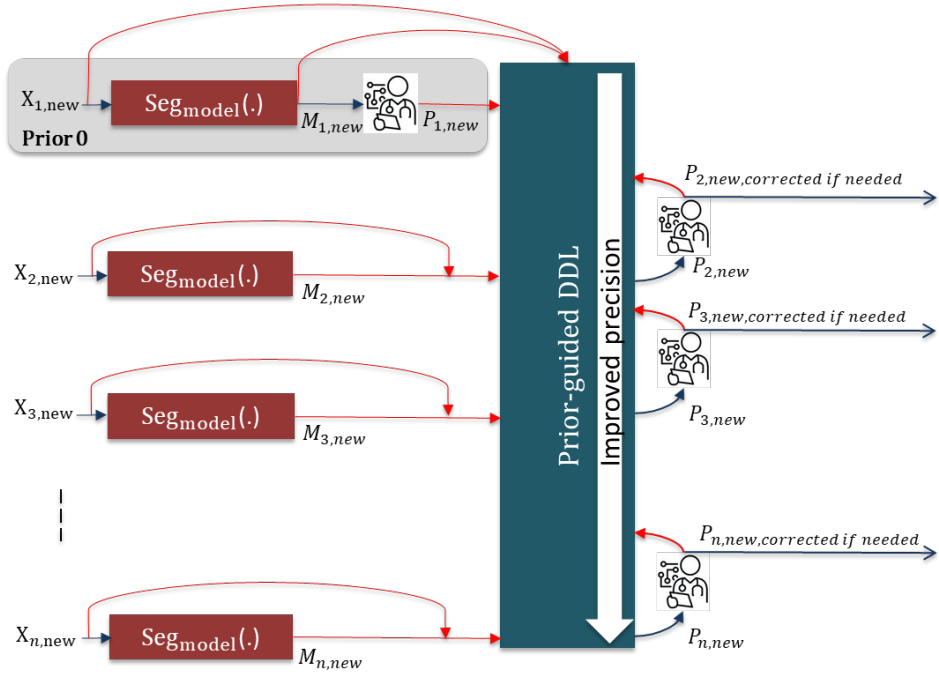

Figure 1: The clinical workflow for using the proposed Prior-guided DDL to a completely new segmentation style.

To the best of our knowledge, this is the first work that addresses the problem of labeling style variability of structures in radiotherapy treatment planning. We provide the first work that uses a meta-learner to gradually and quickly learn the systematic difference between the segmentation label from a pre-trained general auto-segmentation model and the desired special segmentation style as prior-guidance to adapt the label to the new style. The proposed method was tested on three clinical practice style variations: CTV tumor contouring, parotid gland and rectum OAR contouring variations.

## II. METHODS

### Model Architecture

The Prior-guided DDL model has 3 goals: 1) Learn the difference between the pre-trained general segmentation model prediction and the new desired style; 2) The model should be able to predict the new style from prior patients without updating any model parameters ( any further training); and 3) the model should have incremental performance with more prior patients. In this section, we first define the setup of prior guidance, and then we introduce our new Prior-guided DDL model, which learns the difference in styles and produces a stylized segmentation. **Figure 2** illustrates the high-level model diagram of our method. The model has four parts. A pre-trained general segmentation model, which from hereon will be referred to as the segmentation model, a shared encoder, a DDL block and a decoder. The pre-trained segmentation model is the main component of prior-guidance. It generates the initial segmentation $M_n$ from the CT image $X_n$, which needs to be adapted to the new style. The style to be learned is in the form of ground truth segmentation $P_n$ for the same patient.

The pairs $\{X_n, M_n\}$ and $\{X_n, P_n\}$ are input into the same encoder for feature extraction. The encoder is a multi-layer CNN with two Maxpooling layers. Each layer consists of two 3D convolution blocks with ReLU activation followed by GroupNormalization. The final layer consists of three Atrous convolution blocks with dilation rates 6,12 and 24. The resulting encoded features, $f(P_n)$ and $f(M_n)$ for a prior patient are input into the DDL block. The goal of the DDL block is to subtract the initial segmentation and style segmentation in the feature space. DDL Block consists of a difference layer followed by a 3D convolution layer with tanh activation function. If multiple prior patients are available, the output of the tanh layer is averaged for all the prior features. The query data $\{X_{new}, M_{new}\}$ is passed through the same encoder and the resulting features are concatenated with the output of DDL block. These features are then decoded back using a multi-layer CNN to produce the stylized segmentation. Model was trained with a combined DSC and Hausdorff distance loss [39].

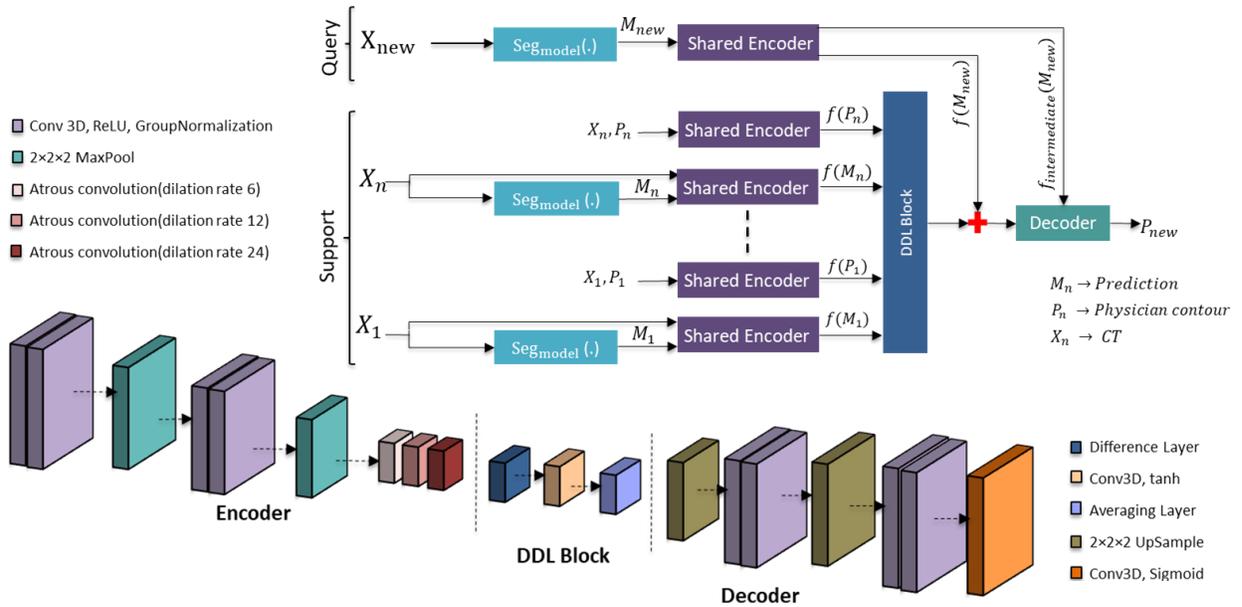

**Figure 2:** Prior-guided DDL framework. Support framework learns the style differences between the segmentation from the pre-trained auto-segmentation model and the new segmentation style. Feature differences are averaged across multiple prior patients, if available, and concatenated with the new patient features. These features are then decoded to get the style-adapted structure.

## Training Details

A combination of DSC loss and Hausdorff distance (HD)-based loss [39] was used for model training. We used an epoch varying weight to balance these two losses such that initially DSC loss has the larger value and once DSC goes above 70%, the weighted HD-based loss had values similar to the DSC loss for any training sample. We used Adam optimizer with $10^{-3}$ learning rate. The batch size was set to one. The model is trained to accept up to 10 prior patients.

The model was trained to accept any number of prior patients from one to ten. At every iteration, the model is exposed to a random number of prior datasets. The Prior-guided DDL model is trained and designed with the assumption that data for the new style is unavailable at the time of training. The meta learner should be capable of learning the difference in style for any new style encountered. To facilitate this, the meta-learner needs to be trained with multiple systematic style variations compared to the original structure. For meta-learner training, new styles were simulated from the available whole structure training data. The augmentations used for different style label simulation is detailed in **Data Augmentation & Simulation** section.

For the segmentation model that produces prior segmentation, a 3D multi-task network was used for CTV [25], a 3D UNet++ [40] was used for parotid segmentation, and a 3D UNet architecture was used for rectum [19].

## Dataset

### Training Data

Post-operative CTV dataset consists of 100 patients treated by a single physician with adjuvant/salvage radiotherapy at UT Southwestern Medical Center (UTSW) from 2017 to 2020. At the UTSW genitourinary (GU) radiation oncology service, these clinical contours are usually drawn by residents, corrected by supervising attending physician, and reviewed by all attending physicians. Each CT volume contains 60-360 slices and a voxel size of $1.17 \times 1.17 \times 2$ mm$^3$.

As for parotid gland training dataset, we used the open access H&N CT scans of patients with nasopharyngeal cancer from the Automatic Structure Segmentation for Radiotherapy Planning Challenge. Each CT scan was marked by one experienced oncologist and verified by another experienced oncologist. All of the original CT scans in this dataset consist of 100~144 slices of 512 × 512 pixels, with a voxel resolution of ($[0.98$~$1.18] \times [0.98$~$1.18] \times 3.00$ mm$^3$).

Whole rectum dataset consists of 220 patients with prostate cancer treated at UTSW from 2017 to 2019. The dataset consisted of raw CT scan images of 136 prostate cancer patients. All CT images were acquired using a 16-slice CT scanner (Royal Philips Electronics, Eindhoven, The Netherlands). Rectum was contoured by experienced radiation oncologists. All images were acquired with a 512x512 matrix and 2 mm slice thickness (voxel size 1.17mm×1.17mm×2mm).

### Testing Data

For post-operative CTV segmentation, 30 patients were used for testing with 3 style variations created by using common variations observed across institutions and different guidelines [33-38]: 1) $CTV_{style1}$ extending only 1cm into Bladder; 2) $CTV_{style2}$ with superior rectum included; and 3) $CTV_{style3}$ excluding the seminal vesicles. Each CT volume contains 60-360 slices and a voxel size of $1.17 \times 1.17 \times 2$ mm$^3$.

For rectum segmentation, two sub-structures were tested with Prior-guided DDL: $Rectum_{superior}$ and $Rectum_{posterior}$. which are the posterior and superior aspects of the rectum respectively. 11 patients with $Rectum_{superior}$ and $Rectum_{posterior}$ contoured by expert physicians were used for testing. All images were acquired with a 512x512 matrix and 2mm slice thickness (voxel size 1.17mm×1.17mm×2mm).

For parotid gland segmentation, $parotid_{superficial}$ is used as the label variation for testing. Superficial parotid gland is just the superficial lobe of the parotid gland without the deep lobe included. 25 patients with superficial parotid contoured by an expert physician at UTSW was used as the test dataset. All the scans in this dataset contain 124~203 slices of $512 \times 512$ pixels, with a voxel resolution of ($[1.17~1.37] \times [1.17~1.37] \times 3.00$ mm$^3$).

### Data Augmentation & Simulation

For training the DDL meta-learner, multiple data augmentations were performed on the whole structure. For CTV dataset, the following data augmentations were performed: structure dilation, structure erosion, CTV completely excluding bladder, CTV 3cm into bladder, CTV 2cm into rectum, CTV contours only in slices with rectum and bladder.

For parotid gland, the data-augmentations used for training the meta-learner included structure dilation, structure erosion, parotid contours only on slices with masseter muscles, parotid contours only on slices with spinal canal and parotid contours only in slices without masseter muscles or spinal canal.

For rectum dataset, the following data augmentations were performed: structure dilation, structure erosion, rectum contours only on slices with bladder, rectum contours only in slices with prostate, rectum contours only in slices without prostate or bladder. These augmentations were chosen so that they are systematic but diverse enough to capture different kinds of variations possible.

## III. EXPERIMENTS & RESULTS

To quantify the performance of our model, we tested the capability of model in learning unseen structure styles. For parotid and rectum, we used existing style variations in the form of $Rectum_{superior}$, $Rectum_{posterior}$ and $Parotid_{superficial}$. For CTV, we simulated three new style variations which are common variations in CTV contouring observed in different institutions and across guidelines. The testing data details are explained in **Testing Data** section. We used DSC for quantifying the results.

We also compared the performance of our model with transfer learning. Transfer learning was performed on the segmentation model with 1-10 patients. Extensive data augmentations in the form of image flip, rotation and x-y-z shift were used for training. The performance of transfer learning model was compared to prior-guided DDL models by comparing the DSC on test data with increasing number of prior patients available.

**Table 1:** Performance of the pre-trained general segmentation model when tested on the whole rectum, parotid gland, and existing CTV style (a) and when tested on the new styles (b)

(a)

| STRUCTURE | DSC(%) [Test dataset] |
|---|---|
| CTV | 85.8±6.80 |
| Parotid Gland | 87.6±3.40 |
| Rectum | 85.4±4.70 |

(b)

| STRUCTURE | DSC(%) [Test dataset] |
|---|---|
| $CTV_{style1}$ | 78.6±6.29 |
| $CTV_{style2}$ | 71.9±7.57 |
| $CTV_{style3}$ | 63.0±9.56 |
| Superficial Parotid Gland | 69.6±12.1 |
| Rectum Posterior | 52.2±12.4 |
| Rectum Superior | 46.3±7.60 |

**Table 1a** shows the performance of the pre-trained general segmentation model on whole rectum, parotid gland, and existing CTV style. The DSC on test data set is above 85% for all three structures. When applying the general model to the new styles to be adapted to, the performance of the model, as tabulated in **Table 1b**, dropped significantly.

**Figure 3** shows the performance of the Prior-guided DDL model on $Parotid_{superficial}$. **Figure 3(a)** The average DSC on the test dataset with the proposed model. The plot shows the improvement in DSC as the number of input prior patients increases. The error bars show the standard deviation. The DSC improvement slows down after 6 prior patients. **Figure 3(b)** shows a visualization of Prior-guided DDL performance for $Parotid_{superficial}$ gland with 4 prior patients as input (green: whole parotid segmentation model prediction, red: ground truth $Parotid_{superficial}$ gland contoured by physician, yellow: Prior-guided DDL stylized contour with 4 input prior patients). With just 4 prior patients as input, it can be observed that the model is able to segment the $Parotid_{superficial}$ much more precisely compared to the whole structure segmentation model.

The performance of the Prior-guided DDL model on rectum labeling variations is visualized in **Figure 4**. **Figure 4(a)** plots the average DSC on the test dataset. The DSC increases as the number of input prior patients increases. For $Rectum_{superior}$ only three prior patients were available for testing. **Figure 4(b)** shows a visualization of model performance for $Rectum_{posterior}$ with 3 prior patients as input (green: whole rectum segmentation model prediction, red: ground truth $Rectum_{posterior}$ contoured by physician, yellow: Prior-guided DDL stylized contour with three input prior patients). **Figure 4(c)** shows a visualization of model performance for $Rectum_{superior}$ with three prior patients as input (green: whole rectum segmentation model prediction, red: ground truth $Rectum_{superior}$ contoured by physician, yellow: Prior-guided DDL stylized contour with three input prior patients). For both $Rectum_{superior}$ and $Rectum_{posterior}$, it can be observed that the DDL stylized contour predicted a much more precise segmentation compared to the whole rectum segmentation model.

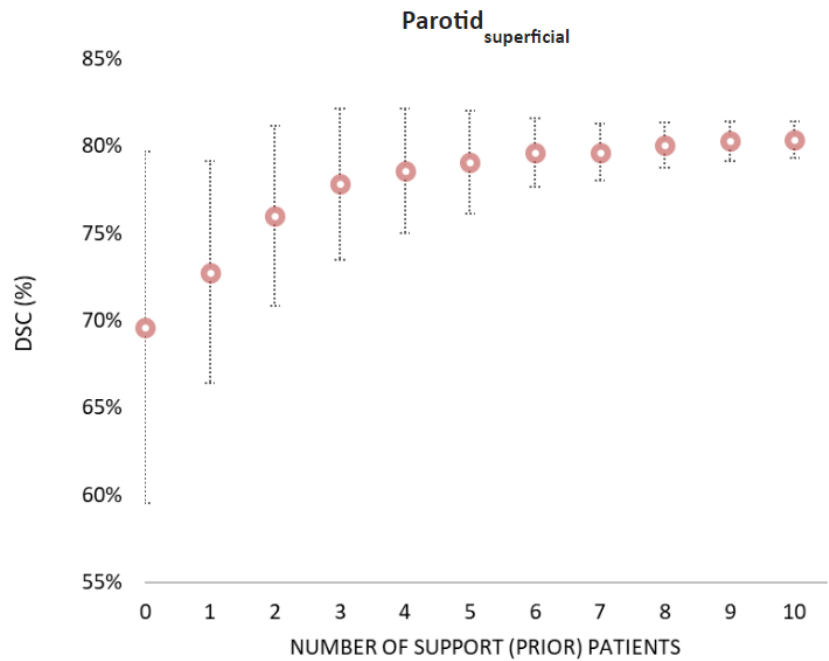

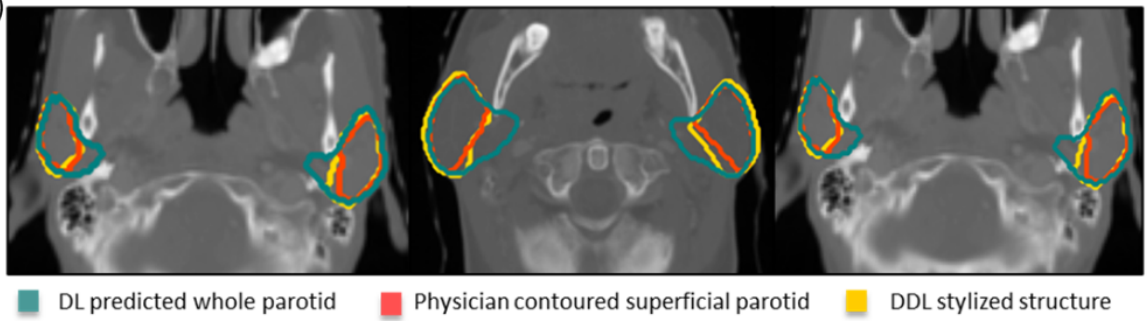

**Figure 3:** Showcase for the segmentation of Parotid$_{superficial}$ gland. (a) The average DSC on the test dataset for DDL model. The plot shows the improvement in DSC as the number of input prior patients increases. The DSC improvement slows down after 6 prior patients. (b) Visualization of DDL performance for Parotid$_{superficial}$ gland with 4 prior patients as input (green: whole Parotid segmentation model prediction, red: ground truth superficial Parotid gland contoured by physician, yellow: DDL stylized contour with 4 input prior patients)

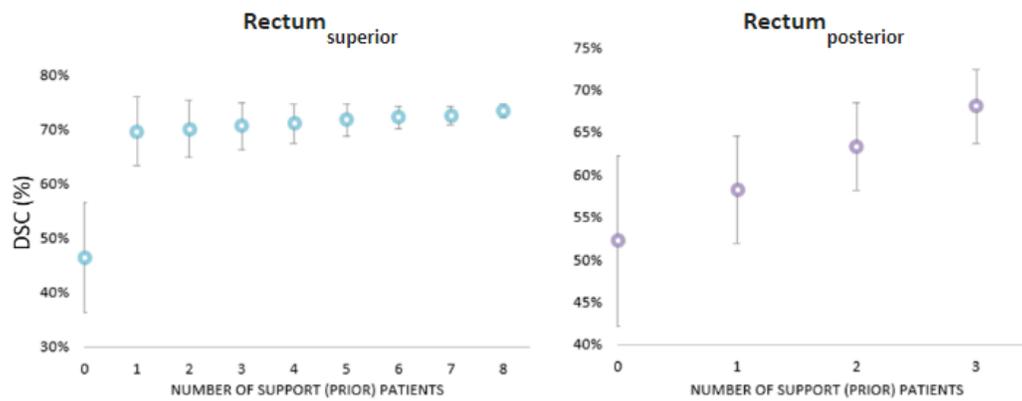

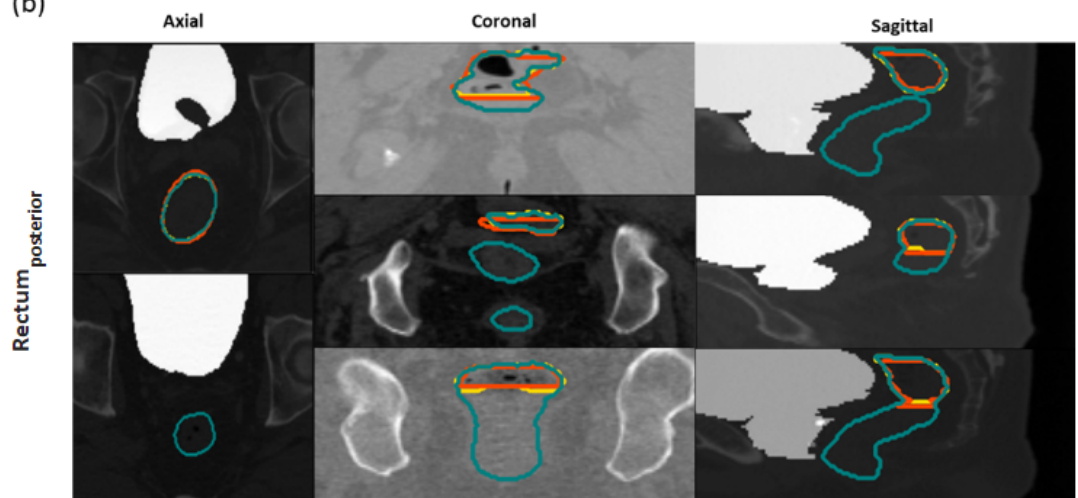

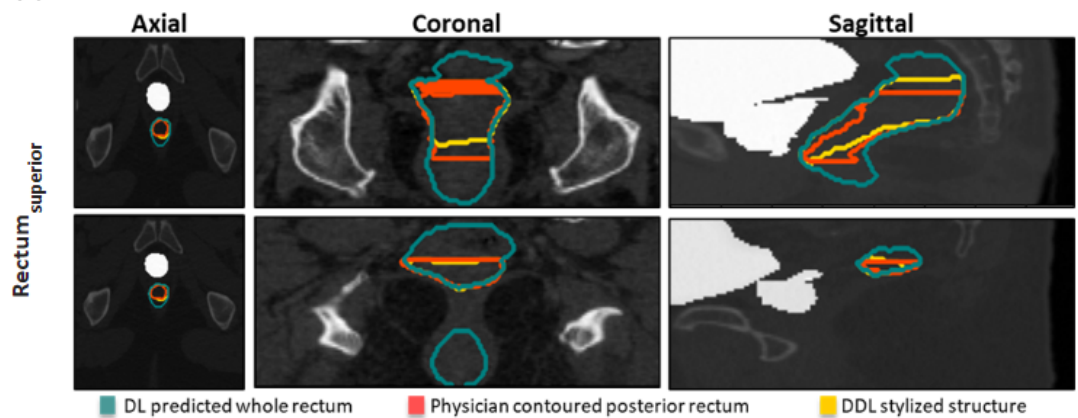

**Figure 4:** Showcase for the segmentation of posterior and Rectum$_{superior}$. (a) The average DSC on the test dataset for DDL model. The plot shows the improvement in DSC as the number of input prior patients increases. (b) Visualization of model performance for Rectum$_{posterior}$ on axial, coronal and sagittal slices with 3 prior patients as input (green: whole rectum segmentation model prediction, red: ground truth Rectum$_{poserior}$ contoured by physician, yellow: DDL stylized contour with 3 input prior patients). (c) Visualization of model performance for Rectum$_{superior}$ on axial, coronal and sagittal slices with 3 prior patients as input (green: whole rectum segmentation model prediction, red: ground truth Rectum$_{superior}$ contoured by physician, yellow: DDL stylized contour with 3 input prior patients).

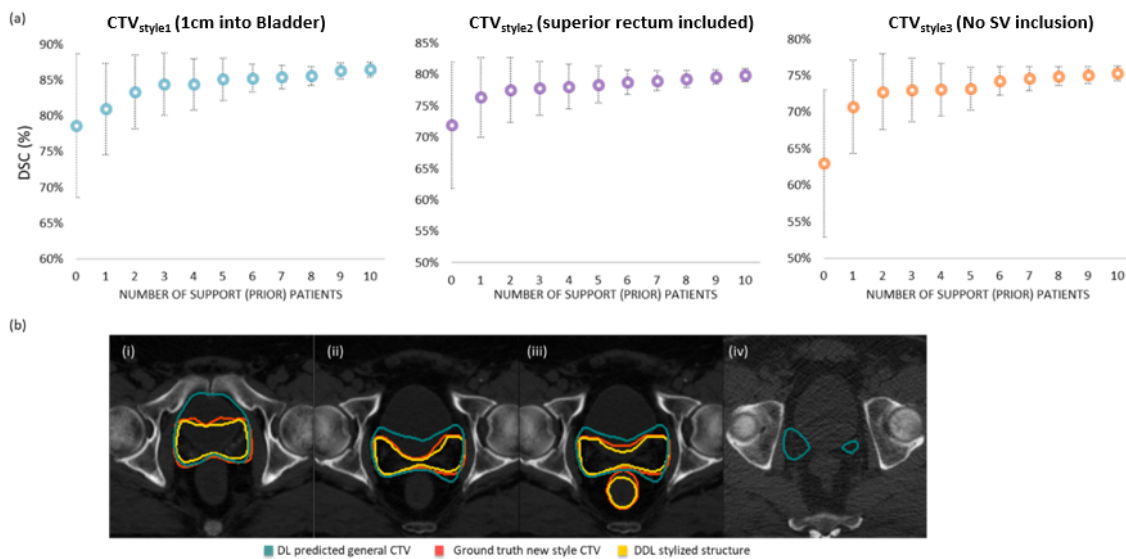

**Figure 5:** Showcase for the segmentation of CTV style variations. (a) The average DSC on the test dataset for Prior-guided DDL model. The plot shows the improvement in DSC as the number of input prior patients increases. (b) Visualization of DDL performance for CTV style variations with 5 prior patients as input (green: initial segmentation model prediction, red: ground truth CTV simulated in a new style, yellow: Prior-guided DDL stylized contour with 5 input prior patients. (i) shows an example for CTV 1cm into Bladder, (ii) shows an example for CTV completely excluding Bladder, (iii) shows an example for CTV with superior rectum included and (iv) shows an example for CTV with seminal vesicles excluded.

**Figure 5** shows the performance of the Prior-guided DDL model on CTV style variations. **Figure 5(a)** is a plot for the average DSC on the test dataset for the proposed model. The plot shows the improvement in DSC as the number of input prior patients increases. **Figure 5(b)** shows a visualization of model performance for the simulated CTV styles with 5 prior patients as input. **Figure 5(b)(i)** shows an example for CTV 1cm into Bladder, **Figure 5(b)(ii)** shows an example for CTV completely excluding Bladder,

**Figure 5(b)(iii)** shows an example for CTV with superior rectum included and **Figure 5(b)(iv)** shows an example for CTV with seminal vesicles excluded for a test patient.

To show the ability of the prior guided DDL model in working with just a handful of patients for improved segmentation, we compare the model with transfer learning. **Figure 6** plots the performance of prior guided DDL model against transfer learning for Parotid$_{superficial}$. X-axis represents the number of priors used in prior-guided DDL model and the number of patients used for tuning the transfer learning model. It can be observed that since the number of patients is low, transfer learning model does not perform as well as prior-guided DDL model.

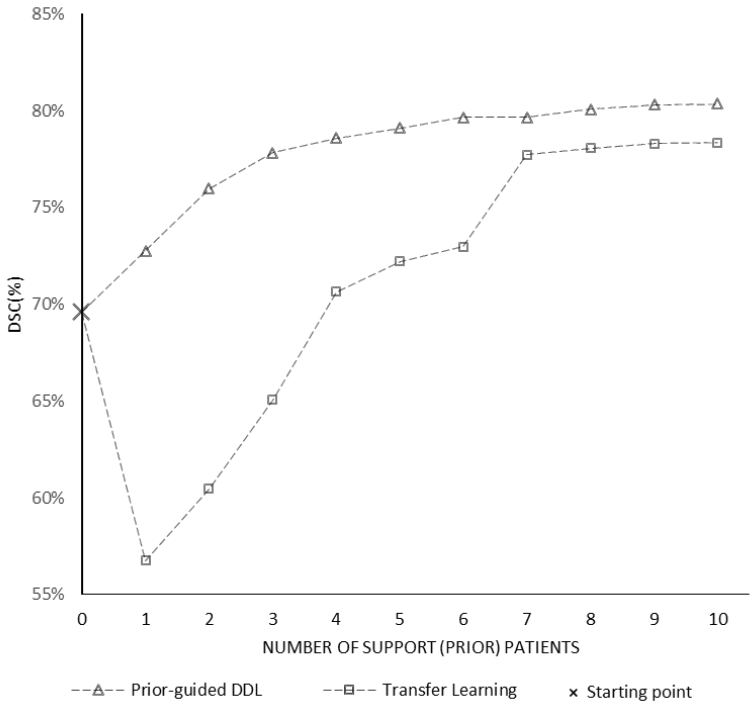

**Figure 6:** Comparison of the prior guided DDL model with transfer learning for model adaptation to Parotid$_{superficial}$. X-axis represents the priors used for prior-guidance in prior-guided DDL model and the number of patients used for tuning the transfer learning model. It can be observed that since the number of patients is low, transfer learning model does not perform as well as prior-guided DDL model.

# IV. DISCUSSION AND CONCLUSIONS

We have proposed a new model capable of adapting to systematic differences in labeling styles across structures with only a handful of patient labels. The proposed model directly addresses the problem of labeling variations that a lot of artificial intelligence (AI) models face when deployed into clinical practice.

AI models for radiotherapy structure segmentation have generalizability issues which could stem from either domain variations or label variations. Physicians utilize the segmented structures to indicate the tumor volumes for treatments as well as the specific areas of OARs to limit the radiation dose to achieve the optimal therapeutic goals. This introduces large labeling differences for the same structures which does not come under the umbrella of either inter-observer or intra-observer variability. A few relevant scenarios include: (1) practice variation in segmenting clinical target volume making it smaller or larger or including/excluding surrounding anatomies; Or, instead of segmenting the whole organ, physicians require only a small part of the organ to be segmented for OAR sparing; (2) as a part of a clinical trial design to change the usual segmentation protocol or would like to treat different parts of the structure differently. Most of these changes are quite systematic with respect to the original structure and having a tool that can learn this difference to produce a more precise "stylized' segmentation would be beneficial to minimize the gap of clinical implementation. A meta-learner allows the users to start from a reasonable baseline contour and painlessly to adapt the model to the desired style that can improve both the quality and the efficiency of the clinical workflow.

In this work we propose a prior-guided deep difference meta-learner that learns how to learn the difference between the segmentation model prediction and a new desired style in order to quickly adapt to the new style. Since the model is meta-trained, it does not have to be trained for a new style and the model can predict the new style from prior patients without updating any model parameters.

To test the proposed approach, we evaluated our model on 7 different structure variations. Experimental results show that, with prior-guided DDL, DSC values improved drastically by 7-20% with just 3 prior patients proving that the model can adapt very well to a new style with just 3 prior patients. The model proposed is capable of learning systematic style differences but will not be able to easily adapt when the style difference is not systematic.

## Acknowledgement

We would like to thank Varian Medical Systems Inc. for supporting this study and Ms. Sepeadeh Radpour for editing the manuscript.

# VI. ACKNOWLEDGMENTS

We would like to thank Dr. Jonathan Feinberg for editing the manuscript.